\pdfoutput=1
\documentclass[preprint,12pt]{elsarticle}

\usepackage{amssymb}
\usepackage{amsmath}
\usepackage{makecell}

\journal{}

\begin{document}

\begin{frontmatter}

\title{A Multimodal Pre-trained Network for Integrated EEG-Video Seizure Detection}

\author[bibr,cibr]{Tong Lu\fnref{equal}}
\ead{lutong@cibr.ac.cn}

\author[sanbo,bibr,cibr]{Ke Xu\fnref{equal}}
\ead{xuke@cibr.ac.cn}

\author[bibr,cibr]{Zimo Zhang}
\ead{zzzzimo.zhang@mail.utoronto.ca}

\author[sanbo,bibr,cibr]{Zitong Zhao}
\ead{zhaozitong_ext@cibr.ac.cn}

\author[genans]{Danwei Weng}
\ead{wengdanwei@genans.cn}

\author[bibr,cibr]{Ruiyu Wang}
\ead{wangruiyu@cibr.ac.cn}

\author[bibr,cibr]{Miao Liu}
\ead{liumiao@cibr.ac.cn}

\author[bibr,cibr]{Zizuo Zhang}
\ead{zhangzizuo@cibr.ac.cn}

\author[bibr,cibr]{Jingyi Yao}
\ead{yaojingyi@cibr.ac.cn}

\author[bibr,cibr]{Yixuan Zhao}
\ead{zhaoyixuan@cibr.ac.cn}

\author[genans]{Wenchao Zhang}
\ead{zhangwenchao@genans.cn}

\author[genans]{Min Wang}
\ead{wangmin@genans.cn}

\author[sanbo,clinmed,epilepsy]{Guoming Luan\corref{cor1}}
\ead{luangm@ccmu.edu.cn}

\author[bibr,cibr,bkl]{Minmin Luo\corref{cor1}}
\ead{luominmin@cibr.ac.cn}

\author[bibr,cibr,bkl]{Zhifeng Yue\corref{cor1}}
\ead{yuezhifeng@cibr.ac.cn}

\affiliation[bibr]{
    organization={Beijing Institute for Brain Research, Chinese Academy of Medical Sciences \& Peking Union Medical College},
    city={Beijing},
    postcode={102206},
    country={China}
}

\affiliation[cibr]{
    organization={Chinese Institute for Brain Research, Beijing},
    city={Beijing},
    postcode={102206},
    country={China}
}

\affiliation[sanbo]{
    organization={Department of Neurosurgery, SanBo Brain Hospital, Capital Medical University},
    city={Beijing},
    postcode={100018},
    country={China}
}

\affiliation[genans]{
    organization={GenAns Biotechnology Co., Ltd},
    city={Beijing},
    postcode={102206},
    country={China}
}

\affiliation[clinmed]{
    organization={Laboratory for Clinical Medicine, Capital Medical University},
    city={Beijing},
    postcode={100069},
    country={China}
}

\affiliation[epilepsy]{
    organization={Center of Epilepsy, Beijing Institute for Brain Disorders, Sanbo Brain Hospital, Capital Medical University},
    city={Beijing},
    postcode={100018},
    country={China}
}

\affiliation[bkl]{
    organization={Beijing Key Laboratory of Brain Science and Brain-Machine Interface},
    city={Beijing},
    postcode={102206},
    country={China}
}

\fntext[equal]{These authors contributed equally to this work.}

\cortext[cor1]{Corresponding authors: Guoming Luan (luangm@ccmu.edu.cn), Minmin Luo (luominmin@cibr.ac.cn), and Zhifeng Yue (yuezhifeng@cibr.ac.cn). Zhifeng Yue is the primary corresponding author.}

%% Abstract
\begin{abstract}
%% Text of abstract
Reliable seizure detection in mouse models is essential for preclinical epilepsy research, yet manual review of synchronized video-EEG recordings is labor-intensive and single-modality systems fail for complementary reasons: video-based methods are easily confounded by benign behaviors, whereas EEG-based methods are vulnerable to ictal motion artifacts. We present EEGVFusion, a multimodal framework that combines self-supervised EEG representation learning, spatio-temporal video encoding, optimal-transport alignment, and bidirectional cross-attention to integrate neural and behavioral evidence. We also curate an expert-annotated dataset of synchronized EEG and video recordings comprising 93 sessions from 15 mice for training and evaluation. In the random-session split, EEGVFusion achieved a Balanced Accuracy of 0.9957 with perfect event sensitivity and an Event FAR of 0.6250 FP/h, indicating strong seizure detection performance with a low false-alarm burden. In a single held-out-subject evaluation with Subject 110 reserved for testing, EEGVFusion achieved a Balanced Accuracy of 0.9718 and reduced Event FAR from 2.7250 FP/h for the EEG-only counterpart to 0.4833 FP/h while preserving perfect event sensitivity. Targeted ablations further showed that EEG pre-training and OT alignment help reduce false alarms while preserving event sensitivity.

\end{abstract}

% %%Graphical abstract
% \begin{graphicalabstract}
% %\includegraphics{grabs}
% \end{graphicalabstract}

% %%Research highlights
% \begin{highlights}
% \item EEGVFusion achieves the strongest overall performance in the random-session split, reaching 0.9957 Balanced Accuracy with perfect event sensitivity and 0.6250 FP/h Event FAR.
% \item We curate and evaluate an expert-annotated dataset containing 93 recording sessions from 15 mice.
% \item Targeted ablations on OT alignment and EEG pre-training clarify the main contributors to robust multimodal seizure detection.
% \end{highlights}

%% Keywords

\end{frontmatter}

\section{Introduction}
\label{sec1}
\begin{figure}[!b]
    \centering
    \begin{minipage}[b]{0.38\textwidth}
        \centering
        \includegraphics[width=\linewidth]{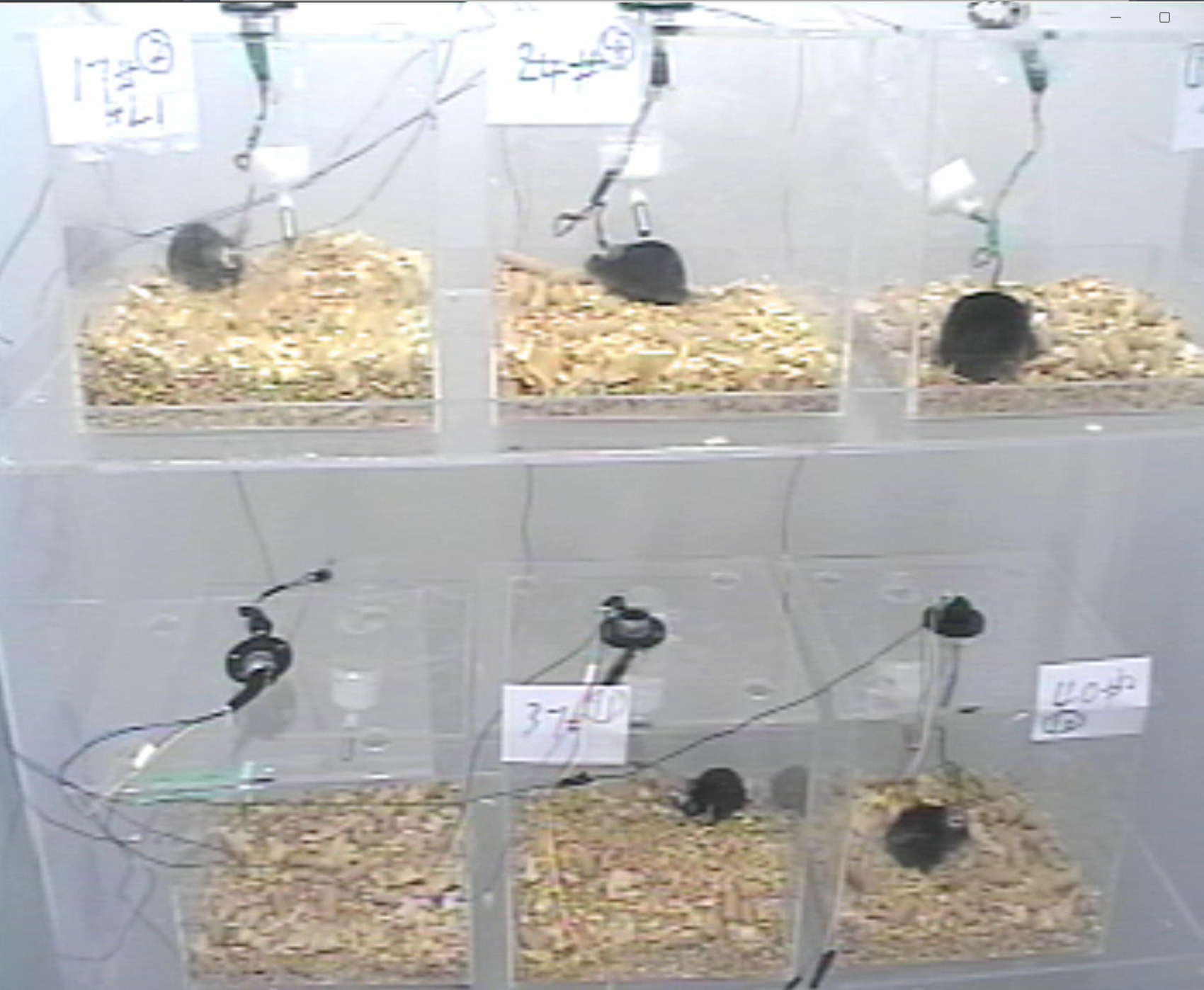}
        \vspace{0.2cm}
        {\footnotesize (a) Behavioral Video Monitoring}
    \end{minipage}
    \hfill
    \begin{minipage}[b]{0.58\textwidth}
        \centering
        \includegraphics[width=\linewidth]{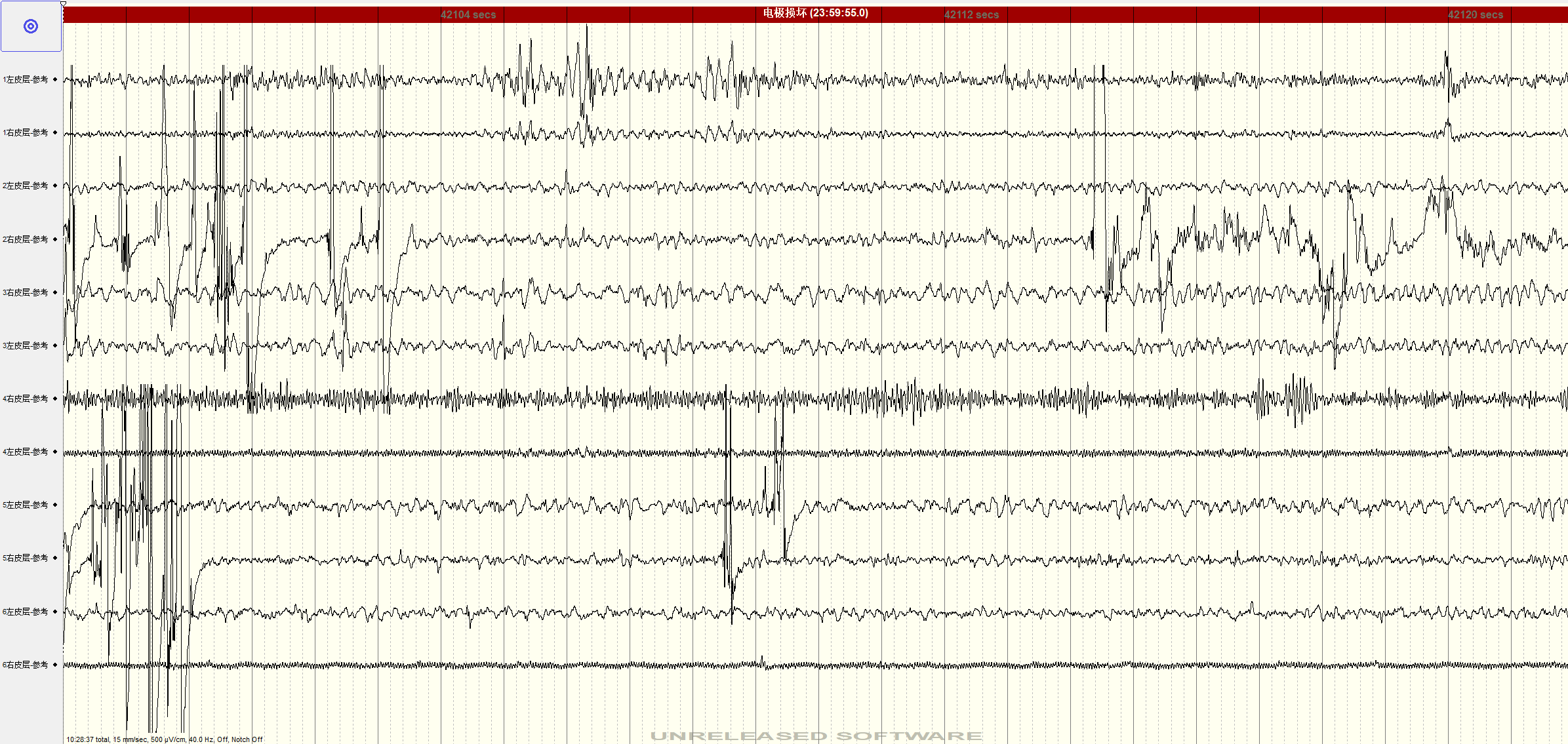}
        \vspace{0.2cm}
        {\footnotesize (b) Synchronized EEG Signals}
    \end{minipage}
    \vspace{0.2cm}
    \caption{Illustration of the high-throughput synchronized continuous monitoring system for mouse epilepsy models. (a) Behavioral video modality capturing macroscopic movements across multiple cages. (b) Synchronized neural EEG modality capturing complex electrophysiological signals. }
    \label{fig:intro_monitoring}
\end{figure}

Epilepsy remains a major global health challenge, affecting more than 50 million people worldwide\cite{laxer2014consequences,beghi2019global}. Rodent, and especially mouse, models are indispensable for studying epileptogenesis and evaluating candidate anti-epileptic therapies during preclinical research\cite{kandratavicius2014animal,edoho2023prediction}. In these studies, synchronized electroencephalography (EEG) and continuous video monitoring remain the reference standard for identifying seizure onset, duration, and behavioral severity. However, annotation still depends largely on manual expert review, making long-term monitoring labor-intensive, difficult to scale, and vulnerable to inter-observer variability\cite{edoho2024ai}. Figure \ref{fig:intro_monitoring} illustrates the complementary data streams that must be interpreted together during routine scoring.

Automating this workflow remains difficult because the two available modalities fail for opposite reasons. Video-based systems are attractive because they are non-invasive and capture overt motor behavior, but vigorous grooming, scratching, or rapid locomotion can closely resemble convulsive seizures and inflate false alarms. EEG-based systems offer direct access to ictal discharges, but they are highly susceptible to motion and muscle artifacts during convulsive events, precisely when reliable detection is most important\cite{raduntz2017automated,fatourechi2007emg,mannan2018effect,tamburro2018new}. These failure modes are complementary: seizure events are often most convincing when behavioral and electrophysiological evidence are evaluated together. The clinical success of multimodal video-EEG systems in human epilepsy supports this view\cite{lin2024vepinet,lin2025development,wu2021multimodal,cao2024synchronized}.

Recent advances in transformer models and self-supervised learning make such multimodal integration increasingly feasible. Self-attention architectures can model long-range temporal dependencies beyond the local receptive fields of conventional convolutional neural networks, and masked autoencoding has emerged as an effective strategy for learning transferable EEG representations from unlabeled recordings\cite{wan2023eegformer,wang2023brainbert,zhang2023brant}. Nevertheless, two obstacles still limit multimodal seizure detection in rodent studies: precisely synchronized mouse EEG-video datasets remain scarce\cite{edoho2024ai}, and current rodent pipelines are still dominated by single-modality analysis without dedicated mechanisms for cross-modal alignment and fusion.

We therefore propose EEGVFusion, a multimodal framework that combines self-supervised EEG representation learning, spatio-temporal video encoding, optimal-transport alignment, and bidirectional cross-attention to improve continuous seizure monitoring in mouse models. Alongside the model, we curate an expert-annotated dataset of synchronized EEG and video recordings to support systematic evaluation of this task. We further conduct targeted ablations on OT alignment and EEG pre-training to clarify which components contribute most to robust detection.

The primary contributions of this work are summarized as follows:

\begin{itemize}

    \item We introduce EEGVFusion, a multimodal seizure-detection framework that integrates self-supervised EEG representations with spatio-temporal video features through optimal-transport alignment and bidirectional cross-attention.
    
    \item We curate a synchronized, expert-annotated mouse EEG-video dataset comprising 93 sessions from 15 mice for training and evaluating multimodal seizure detection models.
    
    \item Empirically, EEGVFusion achieves a Balanced Accuracy of 0.9957 in the random-session split and 0.9718 in a single held-out-subject evaluation, while targeted ablations show that EEG pre-training and OT alignment help control false alarms.

\end{itemize}

\section{Related Work}
\label{sec2}

\subsection{Applications of Transformer Architectures on EEG}

Transformers have substantially improved EEG analysis by modeling long-range temporal and spatial dependencies that are difficult to capture with purely convolutional models. Hybrid architectures such as EEG Conformer\cite{song2022eeg} and EEGformer\cite{wan2023eegformer} combine convolutional front ends with self-attention to represent both local waveform structure and global context. More recent variants extend this idea through specialized attention mechanisms, including criss-cross spatial-temporal modeling\cite{wang2024cbramod} and cross-scale tokenization\cite{zhou2025csbrain}, to better match the heterogeneous structure of EEG signals.

The scarcity of labeled EEG data has also motivated self-supervised foundation models based on masked reconstruction. BrainBERT\cite{wang2023brainbert}, Neuro-BERT\cite{wu2022neuro}, and Brant\cite{zhang2023brant} show that large-scale unlabeled recordings can be used to learn robust and transferable neural representations for downstream EEG analysis. These advances motivate our use of masked autoencoding for EEG feature learning. However, prior transformer-based EEG studies remain largely focused on single-modality representation learning and do not address how synchronized EEG should be aligned with behavioral video for rodent seizure detection.

\subsection{Multimodal Epilepsy Detection in Humans}

Clinical epilepsy monitoring provides strong evidence that multimodal video-EEG analysis can improve robustness when EEG alone becomes unreliable. In pediatric seizure detection, synchronized video features help compensate for motion- and muscle-related distortions in the EEG stream, reducing false alarms while preserving sensitivity\cite{wu2021multimodal,cao2024synchronized}.

Beyond overt seizure detection, multimodal fusion has also proven useful for identifying more subtle epileptic activity. Frameworks such as vEpiNet\cite{lin2024vepinet} and vEpiNetV2\cite{lin2025development} integrate kinematic video features with EEG patterns to improve detection of interictal epileptiform discharges across clinical environments. These studies establish video-EEG fusion as a promising strategy, but they were designed for human clinical recordings with different viewing conditions, annotation practices, and subject variability. Whether the same principles transfer to rodent monitoring remains an open question.

\subsection{Automated Seizure Detection in Rodent Models}

Rodent seizure detection studies have mostly focused on EEG-only automation. Early systems, such as Epi-AI\cite{wei2021detection}, relied on hand-crafted time-frequency features combined with conventional classifiers.

More recent studies have replaced manual feature engineering with deep neural networks. Cho and Jang\cite{cho2020comparison} showed that convolutional architectures outperform recurrent and fully connected alternatives for mouse EEG analysis, and Baser et al.\cite{baser2022automatic} demonstrated effective deep-learning-based detection of spike-and-wave discharges in rats.

Although these methods reduce the burden of manual review, they still process neural signals in isolation. This creates a critical bottleneck during convulsive episodes, when vigorous movement simultaneously produces behavior that is visible in video and artifacts that contaminate EEG. What is still missing is a rodent benchmark with synchronized EEG and video together with a fusion strategy designed to exploit these complementary sources of evidence.

\section{Methodology}
\label{sec3}

\begin{figure}[htbp]
\centering
\includegraphics[width=0.8\textwidth]{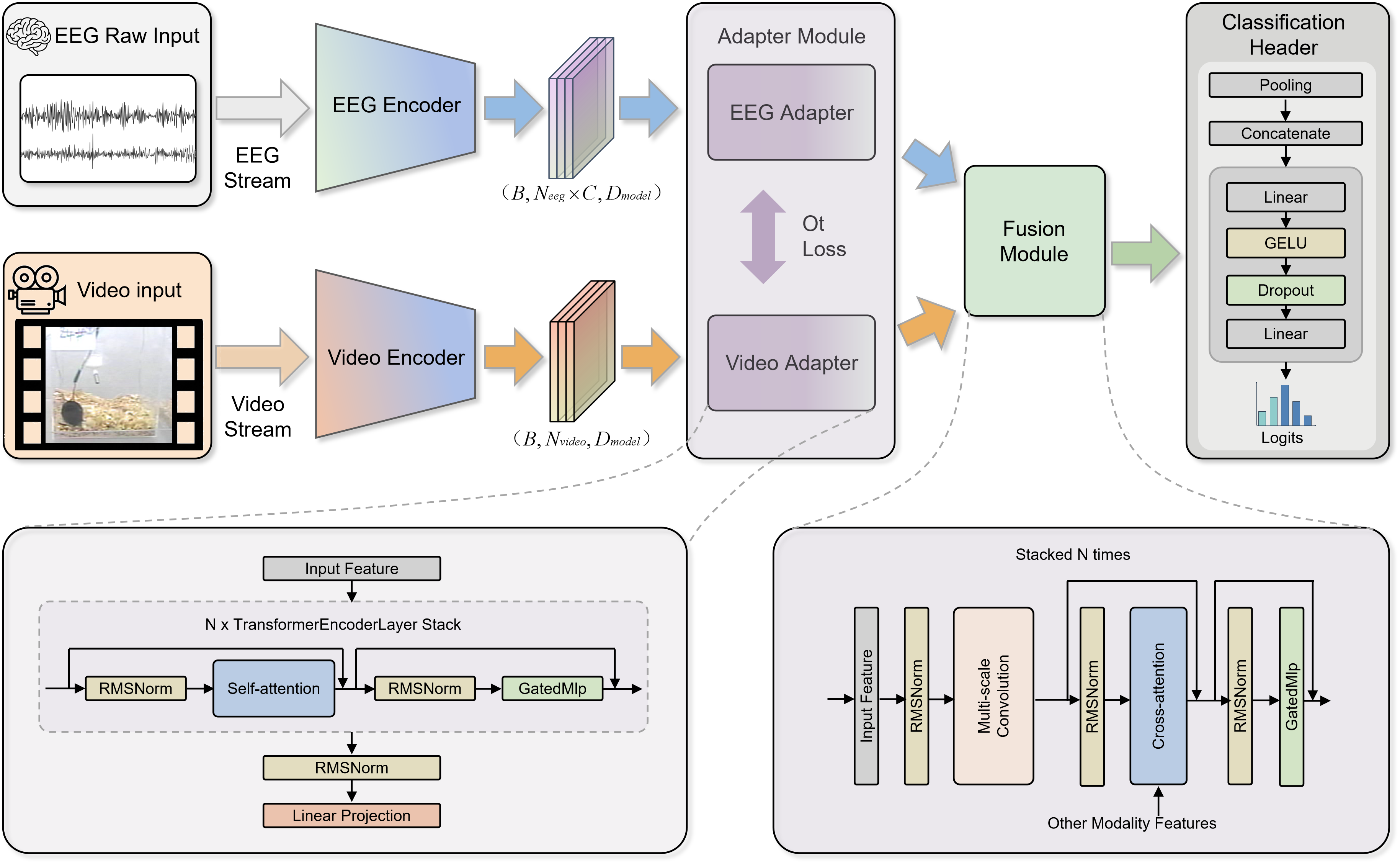}
\caption{Overall architecture of the EEGVFusion framework, illustrating feature extraction, multimodal adaptation, optimal transport alignment, and final classification.}\label{fig:architecture}
\end{figure}

The goal of EEGVFusion is not simply to maximize sample-level classification accuracy, but to improve event-level reliability during long-term monitoring by combining complementary neural and behavioral evidence. The pipeline therefore comprises five stages: signal preprocessing, self-supervised EEG representation learning, spatio-temporal video feature extraction, multimodal alignment and fusion, and temporal aggregation for final event reporting. Figure \ref{fig:architecture} summarizes the full architecture.

\subsection{Signal Acquisition and Preprocessing}

The raw dataset comprises multi-channel EEG signals and synchronized video recordings. Consequently, preprocessing is essential to isolate relevant physiological markers from environmental noise.

\begin{figure}[t]
\centering
\includegraphics[width=1\textwidth]{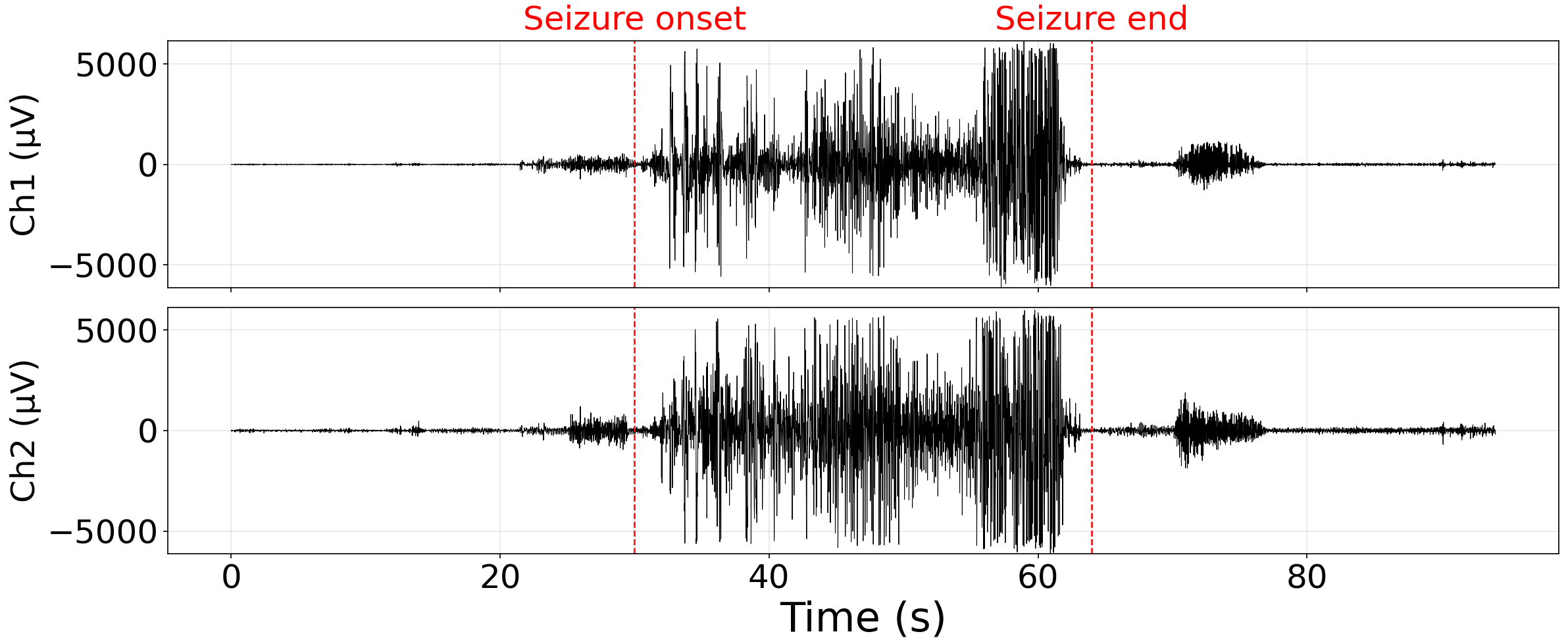}
\caption{Temporal visualization of preprocessed EEG during a seizure. The top and bottom traces correspond to the two recorded channels. The red dashed lines mark seizure onset and termination, which are characterized by high-amplitude rhythmic discharges.}\label{fig:waveform}
\end{figure}

\subsubsection{EEG Preprocessing and Characteristic Analysis}

To mitigate noise while preserving seizure-related activity, the EEG signals undergo digital preprocessing. A 1--50 Hz band-pass filter isolates the frequency range of interest, a 50 Hz notch filter attenuates power-line interference, and baseline wander is corrected using median-filter subtraction.

\begin{figure}[t]
\centering
\begin{minipage}{0.32\textwidth}
  \centering
  \includegraphics[width=\linewidth]{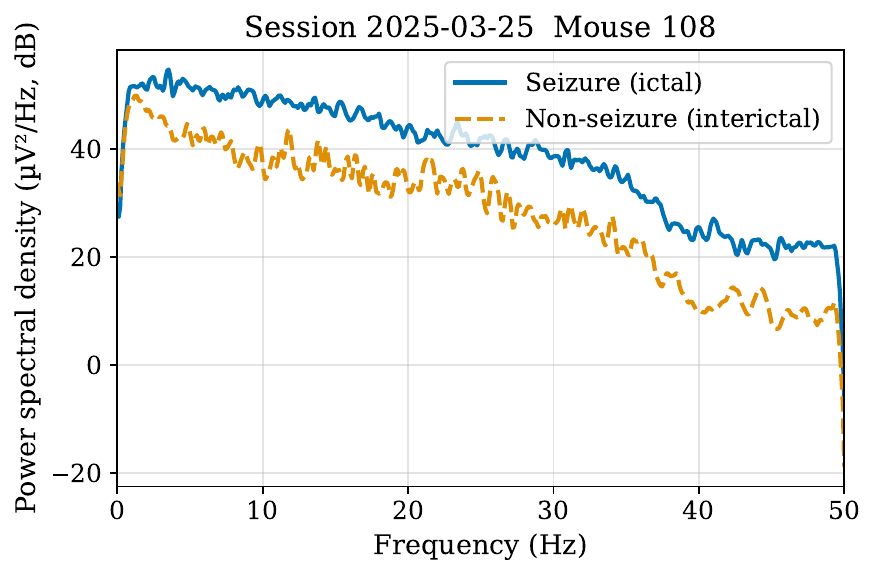}
  \centerline{\footnotesize (a) Mouse 108}
\end{minipage}
\hfill
\begin{minipage}{0.32\textwidth}
  \centering
  \includegraphics[width=\linewidth]{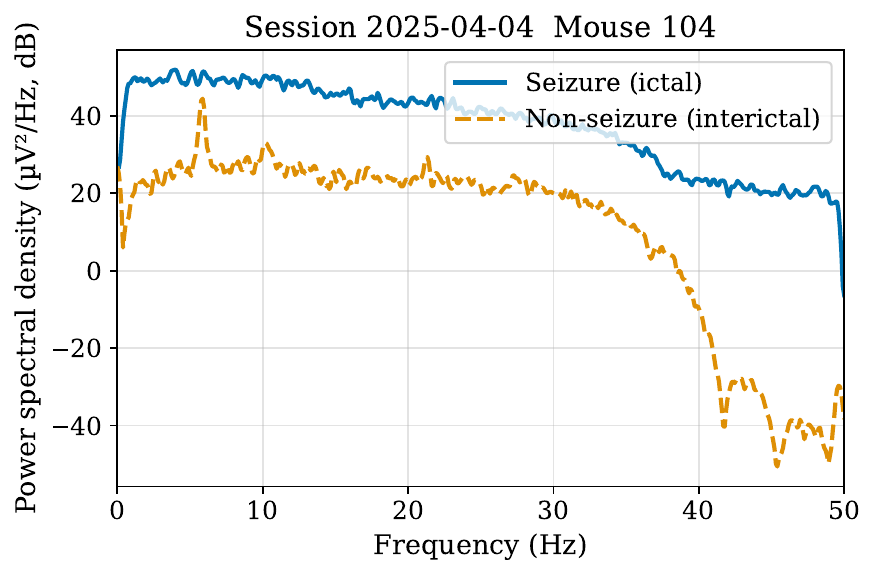}
  \centerline{\footnotesize (b) Mouse 104}
\end{minipage}
\hfill
\begin{minipage}{0.32\textwidth}
  \centering
  \includegraphics[width=\linewidth]{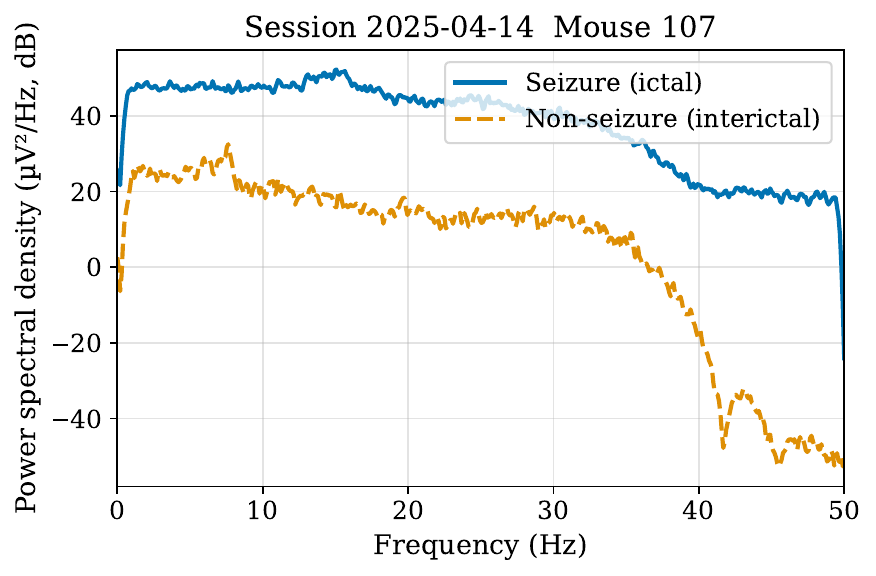}
  \centerline{\footnotesize (c) Mouse 107}
\end{minipage}
\caption{PSD analysis of EEG data from three different mice on different days.}\label{fig:psd}
\end{figure}

As shown in Figure \ref{fig:waveform}, ictal segments contain prominent rhythmic discharges that are clearly separated from the lower-amplitude background activity observed outside seizures. Figure \ref{fig:psd} further demonstrates that seizure and non-seizure periods differ substantially in the frequency domain. Preserving these temporal and spectral signatures during preprocessing is therefore essential for accurate seizure localization.

\subsubsection{Video Preprocessing}

The video stream is originally recorded at 25 fps and temporally downsampled to 6.25 fps by retaining one frame out of every four. This reduces computational redundancy while preserving the macroscopic motor patterns most relevant to seizure detection, including tremor, rearing, and abrupt postural transitions.

\subsection{Self-supervised EEG Feature Representation}

To capture seizure-related EEG structure without depending on extensive labeled data, we employ a Masked Autoencoder (MAE) tailored for neural time series.

\subsubsection{Dual-Domain Embedding}
The continuous EEG signal is segmented into patches before being passed to the transformer encoder. Unlike a standard vision transformer, our encoder uses a dual-domain embedding strategy: temporal features extracted with a 1D convolution preserve local waveform morphology, whereas spectral features obtained through the Fast Fourier Transform (FFT) explicitly encode frequency-domain changes such as those shown in Figure \ref{fig:psd}:

$$\mathbf{E}_{time} = \text{Conv1D}(x) \in \mathbb{R}^{D/2}$$
$$\mathbf{E}_{freq} = \text{Linear}(\text{FFT}(x)) \in \mathbb{R}^{D/2}$$

Here, $D$ denotes the total hidden dimension of the embedding. Learnable positional encodings $\mathbf{E}_{pos}$ are added to the concatenated tokens to preserve temporal ordering and channel context:
$$\mathbf{Z}_0 = [\mathbf{E}_{time}; \mathbf{E}_{freq}] + \mathbf{E}_{pos}$$

\begin{figure}[t]
\centering
\includegraphics[width=1\textwidth]{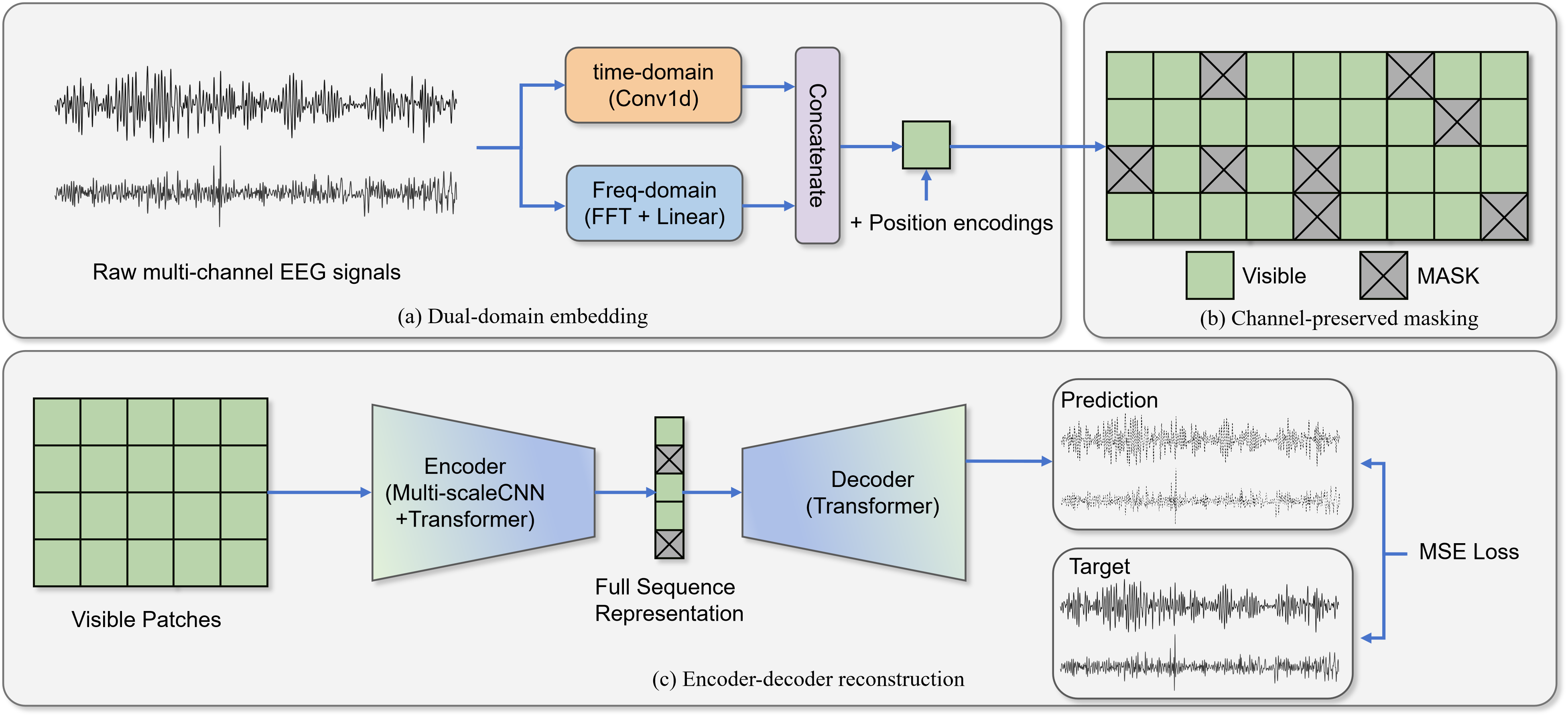}
\caption{The Masked Autoencoder (MAE) pre-training architecture. The model learns robust EEG representations by reconstructing masked patches of the signal in a self-supervised manner.}\label{fig:pretrain}
\end{figure}

\subsubsection{Masked Self-Supervised Learning}

As shown in Figure \ref{fig:pretrain}, the core of our pre-training is a masked reconstruction task. Let $\mathbf{X} \in \mathbb{R}^{C \times N \times P}$ be a multi-channel EEG segment, where $C$, $N$, and $P$ represent the number of channels, patches, and patch length. Using a channel-preserved strategy, we randomly mask a proportion $\rho$ of the patches, dividing the input into visible patches $\mathbf{Z}_{\text{visible}}$ and masked patches $\hat{\mathbf{X}}_{\text{masked}}$ for reconstruction. The encoder, comprising $L_e$ transformer layers, processes only $\mathbf{Z}_{visible}$ to generate contextualized representations $\mathbf{H}_{enc}$. A lightweight decoder then predicts $\hat{\mathbf{X}}_{masked}$ from these representations and learnable mask tokens $\mathbf{M}_{masked}$. The training objective minimizes the mean squared error:

$$\mathcal{L}_{MAE} = \frac{1}{|\mathcal{M}|} \sum_{i \in \mathcal{M}} \|\mathbf{X}_i - \hat{\mathbf{X}}_i\|^2$$

\subsection{Spatio-temporal Video Feature Extraction}

During the ictal phase, mice frequently exhibit erratic movements and may be partially occluded by cage walls or bedding. We therefore employ VideoMAE v2\cite{wang2023videomaev2} for visual feature extraction because its self-attention mechanism is well suited to modeling long-range temporal structure under partial occlusion.

VideoMAE v2 generates a sequence of spatio-temporal tokens that summarizes motion dynamics within a fixed temporal window. This representation is well suited to the rhythmic clonus, tremor, and posture changes that characterize seizure evolution. Because the encoder is pre-trained, the model can extract these patterns without manual bounding-box annotation.

\subsection{Multimodal Adapter and Fusion Framework}

In integrated EEG-video seizure detection, the main fusion challenge is the distributional mismatch between one-dimensional neural signals and spatio-temporal visual tokens. We address this mismatch through modality-specific adapters followed by an Optimal Transport (OT) alignment module, as illustrated in the central pathway of Figure \ref{fig:architecture}.

The adapters project EEG and video features into a shared fusion dimension $D_f$ while preserving modality-specific structure. During multimodal training, the EEG encoder remains frozen so that the pre-trained neural representation is not overwritten by the smaller labeled dataset.

To align the semantic spaces of both modalities, we solve an optimal transport problem over the adapted features $\mathbf{A}_{eeg}$ and $\mathbf{A}_{video}$. A cosine-distance cost matrix $\mathbf{C}$ is used to identify semantically similar cross-modal pairs, and the transport plan $\mathbf{T}^*$ is estimated with the Iterative Proportional Optimal Transport (IPOT) algorithm. The resulting OT distance acts as an alignment loss:

$$\mathcal{L}_{OT} = \lambda_{OT} \cdot \text{tr}(\mathbf{C}^T \mathbf{T}^*)$$

Following alignment, the framework integrates features through a sequence of multi-scale temporal convolutions and bidirectional cross-attention. Multi-scale convolutions first capture local seizure rhythms and short-range temporal context within each modality. Bidirectional cross-attention is then used to let each modality reweight the other according to current signal quality. If EEG is degraded by motion artifact, the model may rely more heavily on video tokens; if video becomes visually ambiguous, EEG can preserve electrophysiological specificity. In this sense, OT and cross-attention are intended to play complementary roles: OT encourages a shared feature geometry, whereas cross-attention enables dynamic evidence exchange between modalities.

\subsection{Classification and Temporal Aggregation}

The framework is optimized in two stages. First, the MAE encoder-decoder is pre-trained on unlabeled EEG data using AdamW. Second, supervised multimodal training is performed with the EEG encoder frozen while the adapters, fusion layers, and classification head are optimized. The final objective combines cross-entropy loss with the OT alignment loss:

$$\mathcal{L}_{total} = \mathcal{L}_{CE} + \mathcal{L}_{OT}$$

During inference, the fused tokens are pooled and passed to a gated MLP classification head consisting of two linear layers with an intermediate GELU activation and dropout. For EEGVFusion, the resulting window-level seizure probabilities are converted to a common 1-second resolution and then post-processed into event-level predictions. Events separated by less than 5 seconds are merged into a single seizure epoch, whereas detections shorter than 10 seconds are discarded to reduce spurious alarms. The same event-merging and minimum-duration rules are used when deriving event-level predictions for the compared models so that Event FAR and event sensitivity are assessed under a common post-processing protocol.

\section{Experiments}
\label{sec4}

\subsection{Dataset Description}

The primary dataset was collected from a mouse model of epilepsy with a hippocampal epileptogenic focus. Each session consisted of continuous synchronized 24-hour video monitoring together with dual-channel EEG recordings. To capture high-fidelity neural signals, custom-fabricated electrodes were implanted in contact with the dura mater, with recording electrodes positioned over the olfactory bulb and the reference and ground electrodes placed over the cerebellum.

Ground truth was established through a two-step annotation protocol. Annotators first reviewed the continuous video recordings to identify behavioral seizure onset and offset times using a modified Racine scale. To reduce subjectivity, subtle behaviors such as facial twitching and head nodding were excluded, and only unambiguous motor manifestations, including tail rigidity, rearing, rearing with bilateral forelimb clonus, falling, and jumping, were labeled as seizure-related behaviors. These candidate intervals were then cross-referenced with the synchronized EEG to determine the corresponding electrophysiological onset and termination times.

The continuous recordings were segmented with a sliding-window strategy using a fixed 10-second window and a 1-second stride. A window received a positive label only when the entire interval fell within the expert-annotated seizure period; otherwise it was assigned a negative label. This conservative labeling rule favors high-confidence event boundaries.

The final dataset comprises 93 sessions collected from 15 mice, where each session denotes one continuous 24-hour recording of synchronized EEG and video. Table \ref{tab:dataset_stats} summarizes the recording duration and seizure distribution and highlights the severe class imbalance characteristic of long-term monitoring data.

\begin{table}[htbp]
    \centering
    \caption{Detailed statistics of the rodent seizure dataset.}
    \label{tab:dataset_stats}
    \small
    \setlength{\tabcolsep}{5pt}
    \begin{tabular}{crrrrr} 
        \hline
        \textbf{Subject ID} & \textbf{Days} & \textbf{\makecell{Total Rec.\\(sec)}} & \textbf{\makecell{Total Rec.\\(hours)}} & \textbf{\makecell{Seizure Dur.\\(sec)}} & \textbf{\makecell{Seizure\\Count}} \\
        \hline
        101 & 5  & 432,000   & 120.00 & 593   & 11  \\
        102 & 4  & 345,608   & 96.00  & 496   & 14  \\
        103 & 1  & 86,400    & 24.00  & 124   & 2   \\
        104 & 8  & 691,548   & 192.10 & 2,897 & 109 \\
        105 & 13 & 1,123,832 & 312.18 & 2,421 & 79  \\
        107 & 5  & 412,383   & 114.55 & 901   & 26  \\
        108 & 7  & 604,802   & 168.00 & 2,271 & 73  \\
        109 & 3  & 259,924   & 72.20  & 221   & 6   \\
        110 & 5  & 432,002   & 120.00 & 706   & 20  \\
        111 & 10 & 805,980   & 223.88 & 1,667 & 41  \\
        112 & 7  & 567,169   & 157.55 & 386   & 12  \\
        113 & 10 & 825,743   & 229.37 & 1,954 & 60  \\
        114 & 4  & 325,983   & 90.55  & 777   & 27  \\
        \#3 & 6  & 519,241   & 144.23 & 658   & 24  \\
        \#4 & 5  & 432,018   & 120.00 & 758   & 24  \\
        \hline
    \end{tabular}
\end{table}

To evaluate both in-distribution performance and robustness to biological variability, we partitioned the data using two complementary protocols:

\begin{itemize}
    \item \textbf{Random Session Split:} Five sessions were randomly selected as the test set and all remaining sessions were used for training. This setting evaluates performance under a standard within-distribution split.
    
    \item \textbf{Single Held-Out-Subject Split:} To obtain an initial test of subject-level generalization, Subject 110 was reserved for testing while the remaining 14 mice were used for training. We interpret this setting as a single held-out-subject evaluation rather than a complete leave-one-subject-out study.
\end{itemize}

As a secondary generalization experiment, we additionally evaluated the fusion design on the Emotion in Audio-Visual (EAV) dataset\cite{lee2024eav}. This multimodal corpus comprises 30-channel EEG, audio, and video recordings from 42 participants during cue-based conversations designed to elicit five emotional states. We followed the preprocessing pipeline of the original study and used this evaluation to examine whether the same fusion strategy transfers beyond seizure detection.

\subsection{Implementation Details and Experimental Setup}

EEGVFusion was implemented in PyTorch and trained on a Linux workstation (Rocky Linux 8.6) equipped with an NVIDIA RTX P6000 GPU. EEG pre-training required approximately 186 hours, and supervised multimodal training for the seizure-detection task required approximately 56 hours.

All experiments used the AdamW optimizer. During pre-training, the learning rate was initialized at $1 \times 10^{-4}$ with a weight decay of 0.05 and a cosine-annealing schedule updated at each step. The model was trained for 1,000 epochs with a batch size of 512 and a masking ratio of 0.75. The asymmetric MAE architecture used an encoder with 8 layers and 8 attention heads and a lightweight decoder with 4 layers and 8 heads; both employed a feed-forward dimension $d_{\mathrm{ff}}=2048$.

During supervised multimodal training, the weight decay was reduced to 0.01. The learning rate again started at $1 \times 10^{-4}$ and decayed to $\eta_{\min}=1 \times 10^{-6}$ under cosine annealing. The model was trained for 30 epochs with a batch size of 64. The fusion module used 4 adapter layers and 4 fusion layers. To mitigate class imbalance, negative samples were downsampled during training to a positive-to-negative ratio of 1:10.

To ensure a fair comparison, all baselines were trained and evaluated on identical dataset partitions. We benchmarked EEGVFusion against established convolutional neural networks (ShallowConvNet, EEGNet, and DeepConvNet\cite{lawhern2018eegnet,schirrmeister2017deep}), a specialized seizure-detection system (Epi-AI\cite{wei2021detection}), a hybrid temporal model (1dCNN-LSTM\cite{kashefi2025epileptic}), and two attention-based architectures (EEGformer\cite{wan2023eegformer} and Husformer\cite{wang2024husformer}).

Because full-length 24-hour recordings exceeded the memory capacity of the RTX P6000 for EEGformer and Husformer, we applied $4\times$ temporal downsampling to the inputs of these two baselines to obtain feasible training runs.

For the secondary EAV evaluation, we compared our approach against the EEG-only, audio-only, and visual-only baselines reported in the original study\cite{lee2024eav}, as well as multimodal emotion-recognition frameworks including AMERL\cite{yin2025eeg} and Hyper-MML\cite{kang2025hypergraph}.

\subsection{Evaluation Metrics}

To evaluate seizure detection comprehensively, we report both sample-level and event-level metrics. Let $TP$, $TN$, $FP$, and $FN$ denote true positives, true negatives, false positives, and false negatives at the sample level, respectively.

\paragraph{Sample-level Metrics}
To account for the varying output granularities employed by different models, we convert every model output together with the ground-truth labels into a uniform \textbf{1-second resolution}. Sample-level evaluation thus assesses each model's ability to correctly classify every 1-second segment.

\begin{itemize}
    \item \textbf{Sample Sensitivity (Recall):} Measures the proportion of actual seizure samples correctly identified.
    $$ \text{Sample Sensitivity} = \frac{TP}{TP + FN} $$
    
    \item \textbf{Sample Specificity:} Measures the proportion of non-seizure (interictal) samples correctly identified.
    $$ \text{Sample Specificity} = \frac{TN}{TN + FP} $$

    \item \textbf{Balanced Accuracy:} Since seizure data is often highly imbalanced (interictal periods are much longer than ictal periods), Balanced Accuracy provides a fairer assessment by averaging sensitivity and specificity.
    $$ \text{Balanced Accuracy} = \frac{\text{Sample Sensitivity} + \text{Sample Specificity}}{2} $$
\end{itemize}

\paragraph{Event-level Metrics}
For long-term preclinical monitoring, correctly identifying seizure events is more important than classifying every isolated second. We therefore complement sample-level metrics with event-level metrics, where an event-level true positive is defined as a predicted seizure that overlaps with a ground-truth seizure period. Unless otherwise stated, event-level detections for all compared models are derived from the common 1-second representation using the same post-processing rules: detections separated by less than 5 seconds are merged, and detections shorter than 10 seconds are removed.

\begin{itemize}

    \item \textbf{Event Sensitivity:} The ratio of correctly detected seizure events to the total number of actual seizure events.

    \item \textbf{Event False Alarm Rate (Event FAR):} To reduce alarm fatigue during long-term preclinical monitoring, FAR quantifies the frequency of false detections over time, calculated as the average number of false seizure detections per hour (FP/h).
    
\end{itemize}

\begin{table}[!t]
\centering
\caption{Comparison of model performance in the random-session split.}
\label{tab:random_session}
\small
\setlength{\tabcolsep}{4.5pt}
\renewcommand{\arraystretch}{1.15}
\hspace*{-1.3cm}
\begin{tabular}{lccccc}
\hline
\textbf{Model Name} 
& \makecell{\textbf{Balanced}\\\textbf{Accuracy}} 
& \makecell{\textbf{Sample}\\\textbf{Sensitivity}} 
& \makecell{\textbf{Sample}\\\textbf{Specificity}} 
& \makecell{\textbf{Event}\\\textbf{Sensitivity}} 
& \makecell{\textbf{Event FAR}\\\textbf{(FP/h)}} \\
\hline
ShallowConvNet~\cite{lawhern2018eegnet} & 0.7114 & 0.4801 & 0.9427 & 0.5533 & 3.8917 \\
EEGNet~\cite{lawhern2018eegnet}         & 0.7829 & 0.6579 & 0.9078 & 0.7000 & 6.3000 \\
DeepConvNet~\cite{lawhern2018eegnet}    & 0.8450 & 0.7196 & 0.9703 & 0.7359 & 2.4167 \\
EEGformer~\cite{wan2023eegformer}      & 0.8709 & 0.9417 & 0.8001 & 0.9667 & 17.0750 \\
Epi-AI~\cite{wei2021detection}        & 0.7610 & 0.7333 & 0.7887 & 0.7333 & 16.7250 \\
1dCNN-LSTM~\cite{kashefi2025epileptic}     & 0.8925 & 0.7994 & \textbf{0.9856} & 0.8000 & \textbf{1.3917} \\
Husformer~\cite{wang2024husformer}       & 0.8542 & 0.7261 & 0.9823 & 0.7533 & 1.5500 \\
\hline
\textbf{EEGVFusion (video-only)}      & 0.9861 & \textbf{1.0000} & 0.9722 & \textbf{1.0000} & 2.4833 \\
\textbf{EEGVFusion (EEG-only)}        & 0.9657 & 0.9595 & 0.9719 & 0.9667 & 2.5167 \\
\textbf{EEGVFusion}  & \textbf{0.9957} & 0.9993 & 0.9921 & \textbf{1.0000} & 0.6250 \\
\hline
\end{tabular}
\end{table}

\begin{figure}[!t]
    \centering
    \begin{minipage}[b]{0.32\textwidth}
        \centering
        \includegraphics[width=\linewidth]{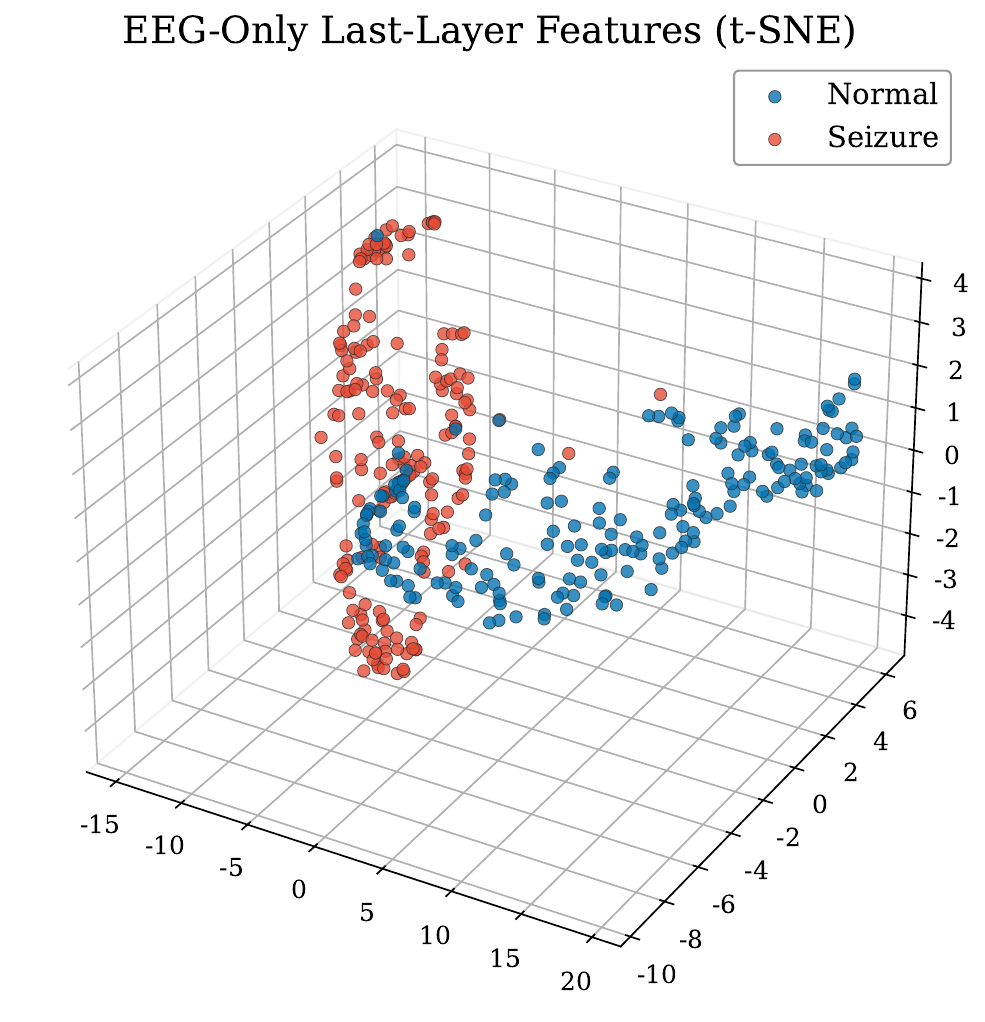}
        \centerline{\footnotesize (a) EEG}
    \end{minipage}
    \hfill
    \begin{minipage}[b]{0.32\textwidth}
        \centering
        \includegraphics[width=\linewidth]{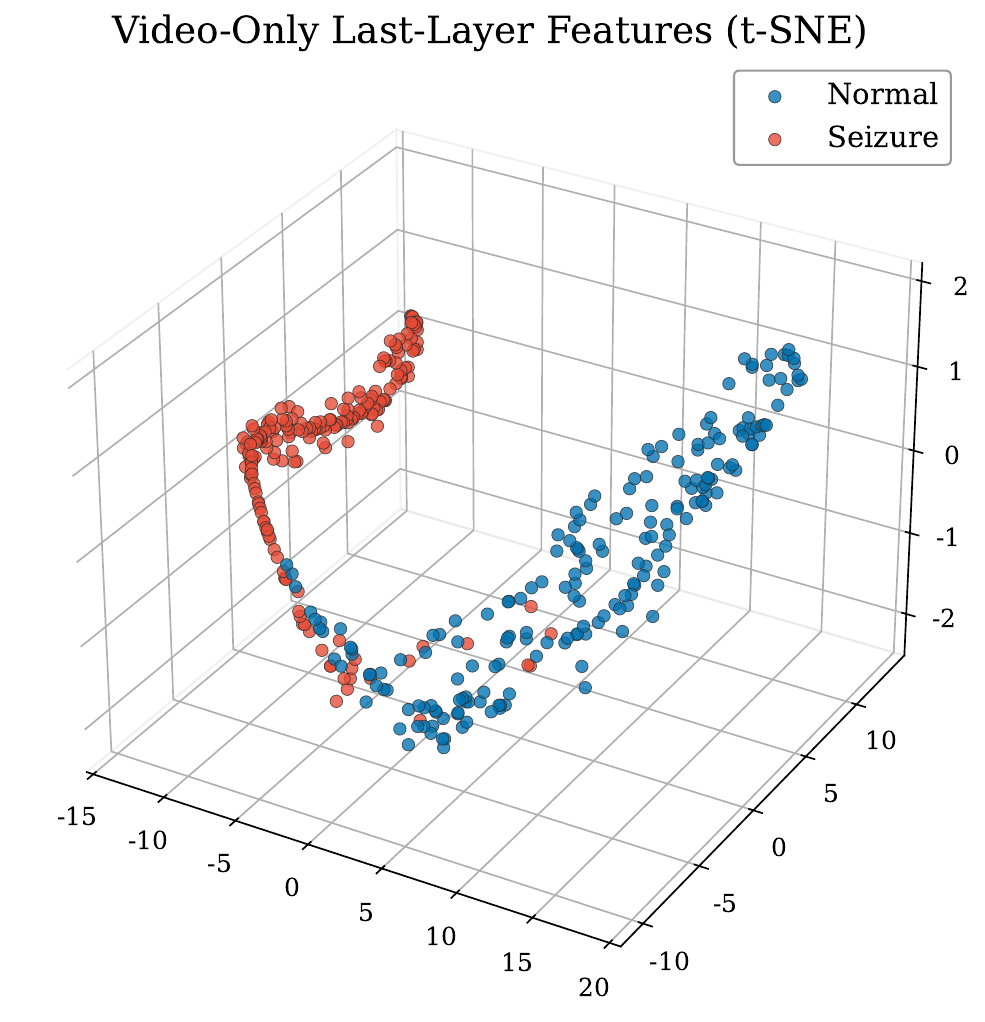}
        \centerline{\footnotesize (b) Video}
    \end{minipage}
    \hfill
    \begin{minipage}[b]{0.32\textwidth}
        \centering
        \includegraphics[width=\linewidth]{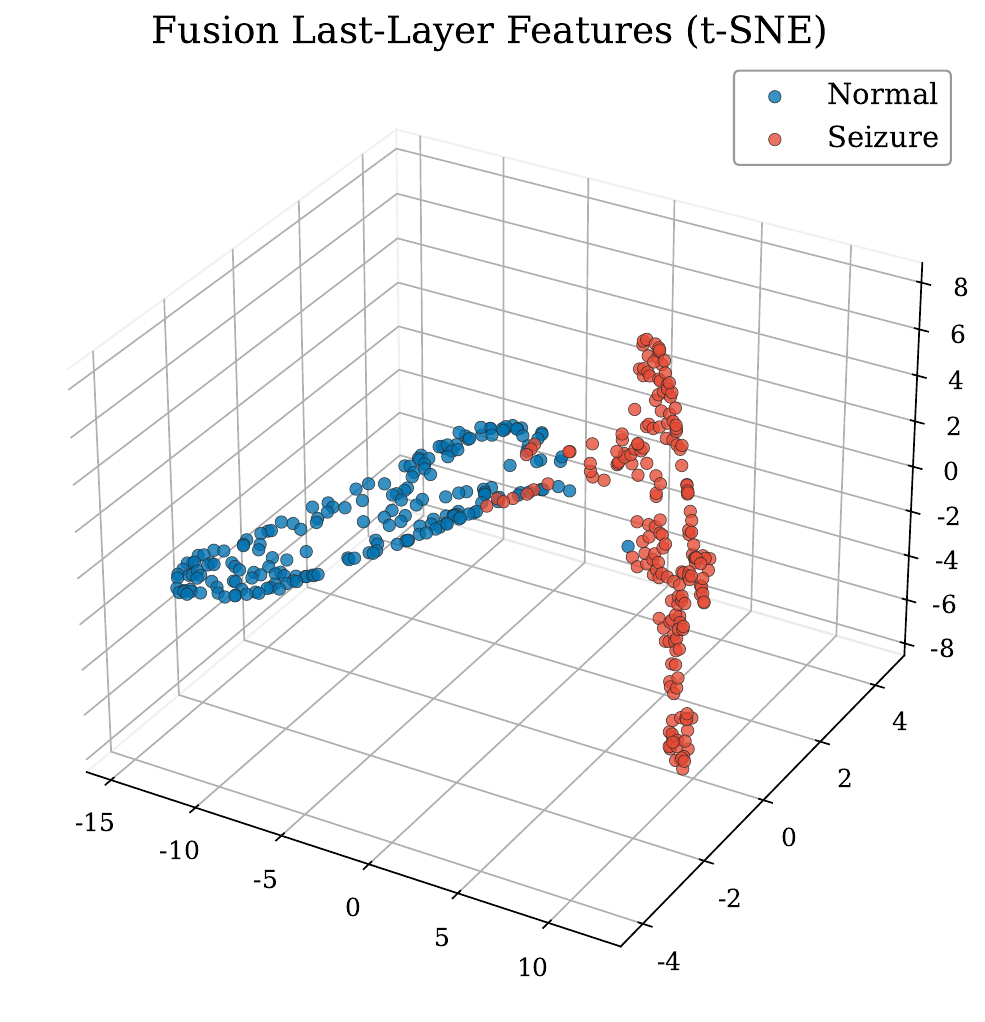}
        \centerline{\footnotesize (c) Fusion}
    \end{minipage}
    \caption{t-SNE visualization of feature distributions: (a) EEG only, (b) Video only, and (c) the proposed Fusion method.}
    \label{fig:tsne_comparison}
\end{figure}

\subsection{Overall Performance}

Tables \ref{tab:random_session} and \ref{tab:cross_subject} summarize performance under the random-session split and the single held-out-subject evaluation. Across both settings, EEGVFusion combines strong balanced discrimination with low false-alarm rates, supporting its use for long-duration preclinical screening.

\subsubsection{Performance in Random Session Split}

In the random-session split (Table \ref{tab:random_session}), EEGVFusion delivered the strongest overall performance, achieving a Balanced Accuracy of 0.9957, a sample sensitivity of 0.9993, perfect event sensitivity, and an Event FAR of 0.6250 FP/h. Compared with 1dCNN-LSTM, a strong and relatively conservative EEG baseline, EEGVFusion improved Balanced Accuracy from 0.8925 to 0.9957 and reduced Event FAR from 1.3917 to 0.6250 FP/h while also raising event sensitivity from 0.8000 to 1.0000. These results indicate that the final multimodal design improves both seizure capture and false-alarm control within distribution. Traditional baselines such as ShallowConvNet and EEGNet remained clearly weaker, with event sensitivities of 0.5533 and 0.7000, respectively.

To verify the feature discriminability, we visualized the last-layer representations using t-SNE on a randomly sampled balanced subset of seizure and non-seizure data. As shown in Figure \ref{fig:tsne_comparison}, the proposed Fusion method (c) achieves a clear separation between seizure and non-seizure clusters, demonstrating that multimodal integration significantly enhances the model's ability to distinguish ictal states.

\begin{table}[t]
\centering
\caption{Comparison of model performance in the held-out-subject evaluation (Subject 110 reserved for testing).}
\label{tab:cross_subject}
\small
\setlength{\tabcolsep}{4.5pt}
\renewcommand{\arraystretch}{1.15}
\hspace*{-1.3cm}
\begin{tabular}{lccccc}
\hline
\textbf{Model Name} 
& \makecell{\textbf{Balanced}\\\textbf{Accuracy}} 
& \makecell{\textbf{Sample}\\\textbf{Sensitivity}} 
& \makecell{\textbf{Sample}\\\textbf{Specificity}} 
& \makecell{\textbf{Event}\\\textbf{Sensitivity}} 
& \makecell{\textbf{Event FAR}\\\textbf{(FP/h)}} \\
\hline
ShallowConvNet~\cite{lawhern2018eegnet} & 0.6679 & 0.3533 & 0.9824 & 0.4167 & 1.6083 \\
EEGNet~\cite{lawhern2018eegnet}         & 0.7741 & 0.5614 & 0.9868 & 0.6300 & 1.2500 \\
DeepConvNet~\cite{lawhern2018eegnet}    & 0.8835 & 0.7951 & 0.9720 & 0.9100 & 2.1917 \\
EEGformer~\cite{wan2023eegformer}      & 0.9158 & 0.9128 & 0.9189 & \textbf{1.0000} & 7.9333 \\
Epi-AI~\cite{wei2021detection}        & 0.9335 & \textbf{0.9923} & 0.8748 & \textbf{1.0000} & 1.2833 \\
1dCNN-LSTM~\cite{kashefi2025epileptic}     & 0.9708 & 0.9513 & \textbf{0.9902} & \textbf{1.0000} & \textbf{0.6250} \\
Husformer~\cite{wang2024husformer}       & 0.8981 & 0.8108 & 0.9857 & 0.8333 & 1.3417 \\
\hline
\textbf{EEGVFusion (video-only)}      & 0.9661 & 0.9454 & 0.9868 & \textbf{1.0000} & 1.1917 \\
\textbf{EEGVFusion (EEG-only)}        & \textbf{0.9739} & 0.9785 & 0.9693 & \textbf{1.0000} & 2.7250 \\
\textbf{EEGVFusion}  & 0.9718 & 0.9493 & 0.9942 & \textbf{1.0000} & 0.4833 \\
\hline
\end{tabular}
\end{table}

\subsubsection{Performance in a Single Held-Out-Subject Evaluation}

The single held-out-subject evaluation (Table \ref{tab:cross_subject}) is substantially more challenging because it tests the model on an animal whose seizure expression and background behavior were not seen during training.

Among our three variants, the EEG-only model achieved the highest Balanced Accuracy (0.9739), indicating that the pre-trained EEG branch remains the strongest individual modality in this held-out-subject setting. The video-only model also remained competitive (0.9661), and both variants outperformed several standard baselines, suggesting that the pre-trained encoders provide a strong starting point even before multimodal fusion is applied.

The event-level picture in this setting is now much clearer. EEGVFusion achieved the lowest Event FAR among all compared methods (0.4833 FP/h) while preserving perfect event sensitivity and maintaining a Balanced Accuracy of 0.9718. Relative to the EEG-only branch, multimodal fusion reduced Event FAR from 2.7250 to 0.4833 FP/h, corresponding to an 82.3\% reduction in false alarms. It also improved upon the strong external baseline 1dCNN-LSTM in both Event FAR (0.6250 vs. 0.4833 FP/h) and Balanced Accuracy (0.9708 vs. 0.9718). These results indicate that multimodal integration can improve practical screening reliability under subject shift, even though broader subject-level validation is still needed.

\subsubsection{Targeted Ablation on OT Alignment and EEG Pre-training}

To clarify the roles of explicit cross-modal alignment and self-supervised representation learning, Table \ref{tab:ablation_key} compares the final EEGVFusion model with two targeted ablations: a no-OT variant and a model trained without EEG pre-training. Removing EEG pre-training reduced Balanced Accuracy and increased the false-alarm burden in the random-session split, while the effect was smaller in the held-out-subject setting. Removing OT preserved perfect event sensitivity but increased Event FAR in both settings, indicating that explicit alignment primarily improves false-alarm control. We therefore interpret EEG pre-training and OT alignment as complementary design choices: pre-training improves representation robustness, whereas OT helps stabilize cross-modal evidence integration.

\begin{table}[t]
\centering
\caption{Targeted ablation of OT alignment and EEG pre-training under the random-session and held-out-subject settings.}
\label{tab:ablation_key}
\small
\setlength{\tabcolsep}{4.5pt}
\renewcommand{\arraystretch}{1.15}
\hspace*{-1.3cm}
\begin{tabular}{llccccc}
\hline
\textbf{Setup} & \textbf{Variant}
& \makecell{\textbf{Balanced}\\\textbf{Accuracy}}
& \makecell{\textbf{Sample}\\\textbf{Sensitivity}}
& \makecell{\textbf{Sample}\\\textbf{Specificity}}
& \makecell{\textbf{Event}\\\textbf{Sensitivity}}
& \makecell{\textbf{Event FAR}\\\textbf{(FP/h)}} \\
\hline
Random session & EEGVFusion & 0.9957 & 0.9993 & 0.9921 & 1.0000 & 0.6250 \\
Random session & No OT & 0.9917 & 1.0000 & 0.9834 & 1.0000 & 1.3000 \\
Random session & No Pretrain & 0.9839 & 0.9845 & 0.9834 & 1.0000 & 1.4750 \\
\hline
Held-out subject & EEGVFusion & 0.9718 & 0.9493 & 0.9942 & 1.0000 & 0.4833 \\
Held-out subject & No OT & 0.9705 & 0.9533 & 0.9878 & 1.0000 & 1.0333 \\
Held-out subject & No Pretrain & 0.9702 & 0.9481 & 0.9924 & 1.0000 & 0.6250 \\
\hline
\end{tabular}
\end{table}

\subsubsection{Secondary Evaluation on the EAV Dataset}

To examine whether the fusion strategy transfers beyond seizure detection, we additionally evaluated EEGVFusion on the Emotion in Audio-Visual (EAV) dataset\cite{lee2024eav}. As shown in Table \ref{tab:eav_results}, EEGVFusion achieved an average accuracy of 0.8139, outperforming the EEG-only, audio-only, and visual-only baselines as well as recent multimodal models including AMERL, Husformer, and Hyper-MML. We treat this result as supporting evidence rather than as a co-primary claim of the paper, but it suggests that the proposed fusion design remains effective on a different multimodal behavior-analysis task.

\begin{table}[t]
\centering
\caption{Comparison of average accuracy on the EAV emotion recognition task.}
\label{tab:eav_results}
\small
\begin{tabular}{lc}
\hline
\textbf{Model Type} & \textbf{Average Accuracy} \\
\hline
EEG-only\cite{lee2024eav} & 0.6000 \\
Audio-only\cite{lee2024eav} & 0.6190 \\
Visual-only\cite{lee2024eav} & 0.7140 \\
AMERL\cite{yin2025eeg} & 0.7086 \\
Husformer\cite{wang2024husformer} & 0.7282 \\
Hyper-MML\cite{kang2025hypergraph} & 0.7821 \\
\textbf{EEGVFusion adaptation} & \textbf{0.8139} \\
\hline
\end{tabular}
\end{table}

\section{Discussion}
\label{sec5}

\subsection{Complementary Modalities}
The main advantage of multimodal fusion is that it addresses the complementary weaknesses of video and EEG. Within the random-session split, this complementarity translated into the strongest overall performance: EEGVFusion achieved the highest Balanced Accuracy, preserved perfect event sensitivity, and reduced Event FAR below that of the strong 1dCNN-LSTM baseline. In other words, the final model no longer trades overall detection quality against false-alarm suppression; it improves both simultaneously within distribution.

The single held-out-subject evaluation reveals an equally important benefit under subject shift. EEGVFusion achieved the lowest Event FAR among all compared methods while preserving perfect event sensitivity. Relative to the EEG-only branch, multimodal integration reduced false alarms from 2.7250 to 0.4833 FP/h, an 82.3\% reduction. For long-term monitoring, this reduction in false events is highly valuable because it substantially decreases the number of segments that must be manually reviewed.

\subsection{Architectural Effectiveness}
Targeted ablation shows that self-supervised EEG pre-training provides a split-dependent benefit. In the random-session split, removing pre-training reduced Balanced Accuracy from 0.9957 to 0.9839 and increased Event FAR from 0.6250 to 1.4750 FP/h. In the held-out-subject evaluation, the no-pretraining variant remained close to EEGVFusion in Balanced Accuracy (0.9702 vs. 0.9718) and preserved perfect event sensitivity, but its Event FAR increased from 0.4833 to 0.6250 FP/h. These results suggest that masked autoencoding most clearly improves in-distribution reliability and still helps false-alarm control under subject shift.

The OT ablation highlights the importance of explicit alignment for false-alarm control. Removing OT reduced Balanced Accuracy from 0.9957 to 0.9917 in the random-session split and increased Event FAR from 0.6250 to 1.3000 FP/h. Under subject shift, the Balanced Accuracy difference was small (0.9705 vs. 0.9718), but Event FAR increased from 0.4833 to 1.0333 FP/h. Thus, OT alignment does not primarily affect whether seizure events are detected, since event sensitivity remained perfect, but it helps suppress false detections during long-duration monitoring.

The secondary EAV result provides additional supporting evidence that the overall fusion strategy is not restricted to rodent seizure monitoring. Achieving 0.8139 average accuracy on a multimodal emotion-recognition task suggests that the combination of modality-specific encoding, explicit alignment, and cross-modal interaction can remain useful in a different behavior-analysis setting. At the same time, we treat this finding as a transferability signal rather than as a central claim of the present study.

\subsection{Limitations}
Despite these strengths, several limitations remain. First, the dataset is severely imbalanced: seizure activity occupies less than 0.25\% of the total recording time. Although negative sampling helps during training, additional evaluation on rare seizure presentations is still needed. Second, this study focuses on a specific hippocampal epilepsy model with only two EEG channels. Future work should test whether the framework generalizes to other rodent models, electrode configurations, and seizure origins. Third, subject-level generalization is currently assessed with a single held-out-subject evaluation rather than a complete leave-one-subject-out study, so stronger claims about robustness across animals require broader validation. Fourth, although targeted ablations now clarify the roles of OT alignment and EEG pre-training, broader analysis of alignment strength, fusion variants, and multi-subject validation is still needed.

\section{Conclusion}
\label{sec6}

We present EEGVFusion, a multimodal framework that combines self-supervised EEG representations with spatio-temporal video features to improve seizure detection in mouse models. In the random-session split, EEGVFusion achieved the strongest overall performance, reaching a Balanced Accuracy of 0.9957 with perfect event sensitivity and an Event FAR of 0.6250 FP/h. In a single held-out-subject evaluation, EEGVFusion achieved a Balanced Accuracy of 0.9718 and reduced Event FAR from 2.7250 to 0.4833 FP/h relative to the EEG-only variant while preserving perfect event sensitivity. Targeted ablations further showed that EEG pre-training and OT alignment help reduce false alarms while preserving event sensitivity. A secondary EAV evaluation achieved an average accuracy of 0.8139, providing supporting evidence that the fusion strategy may transfer beyond seizure detection. Future work should extend subject-level validation, further refine cross-modal alignment, and explore real-time deployment across broader rodent cohorts and recording settings.

\bibliographystyle{elsarticle-num} 
\bibliography{cas-refs}

\end{document}